\pdfoutput=1

\documentclass[11pt]{article}

\usepackage[]{acl}

\usepackage{times}
\usepackage{latexsym}

\usepackage[T1]{fontenc}

\usepackage[utf8]{inputenc}

\usepackage{microtype}
\usepackage{times}
\usepackage{latexsym}
\usepackage{xspace}
\usepackage{booktabs}
\usepackage{multirow}
\usepackage{makecell}
\usepackage{color}
\usepackage{microtype}
\usepackage{amsmath}
\usepackage{graphicx}
\usepackage{amssymb,amsthm}
\usepackage{xcolor}
\usepackage{subcaption}
\usepackage{caption}
\usepackage{paralist}
\usepackage{cleveref}

\usepackage{enumitem}

\usepackage[normalem]{ulem}
\useunder{\uline}{\ul}{}

\newcommand{\our}{\mbox{\textsc{UCTopic}}\xspace}

\title{UCTopic: Unsupervised Contrastive Learning for Phrase Representations and Topic Mining}

\author{Jiacheng Li, Jingbo Shang, Julian McAuley\\
  University of California, San Diego \\
  \texttt{\{j9li,jshang,jmcauley\}@eng.ucsd.edu}\\
}

\begin{document}
\maketitle
\begin{abstract}


High-quality phrase representations are essential to finding topics and related terms in documents (a.k.a. topic mining).
Existing phrase representation learning methods either simply combine unigram representations in a context-free manner or rely on extensive annotations to learn context-aware knowledge.
In this paper, we propose \our, a novel unsupervised contrastive learning framework for context-aware phrase representations and topic mining.
\our is pretrained in a large scale to distinguish if the contexts of two phrase mentions have the same semantics. The key to pretraining is positive pair construction from our phrase-oriented assumptions.
However, we find traditional in-batch negatives cause performance decay when finetuning on a dataset with small topic numbers. Hence, we propose cluster-assisted contrastive learning (CCL) which largely reduces noisy negatives by selecting negatives from clusters and further improves phrase representations for topics accordingly.
\our outperforms the state-of-the-art phrase representation model by $38.2\%$ NMI in average on four entity clustering tasks. 
Comprehensive evaluation on topic mining shows that \our can extract coherent and diverse topical phrases.
\end{abstract}
\section{Introduction}
\label{intro}
Topic modeling 
discovers abstract 'topics' in a collection of documents. A topic is typically modeled as a distribution over terms. 
High-quality phrase representations help topic models understand phrase semantics 
in order to
find well-separated topics and extract coherent phrases. Some phrase representation methods~\cite{Wang2021PhraseBERTIP, Yu2015LearningCM, Zhou2017LearningPE} learn context-free representations by unigram embedding combination. Context-free representations tend to extract similar phrases mentions (e.g.~``great food'' and ``good food'', see~\Cref{sec:topic}). Context-aware methods such as DensePhrase~\cite{Lee2021LearningDR} and LUKE~\cite{Yamada2020LUKEDC} need supervision from task-specific datasets or distant annotations with knowledge bases. Manual or distant supervision limits the ability to represent out-of-vocabulary phrases especially for domain-specific datasets. Recently, contrastive learning has shown effectiveness for unsupervised representation learning in visual~\cite{Chen2020ASF} and textual~\cite{Gao2021SimCSESC} domains.

In this work, we seek to advance state-of-the-art phrase representation methods and demonstrate that 
a
contrastive objective can be extremely effective 
at learning
phrase semantics in sentences. We present \our, an \underline{U}nsupervised \underline{C}ontrastive learning framework for phrase representations and \underline{TOPIC} mining, which can produce superior phrase embeddings and have topic-specific finetuning for topic mining.
\begin{figure}
    \centering
    \includegraphics[width=\linewidth]{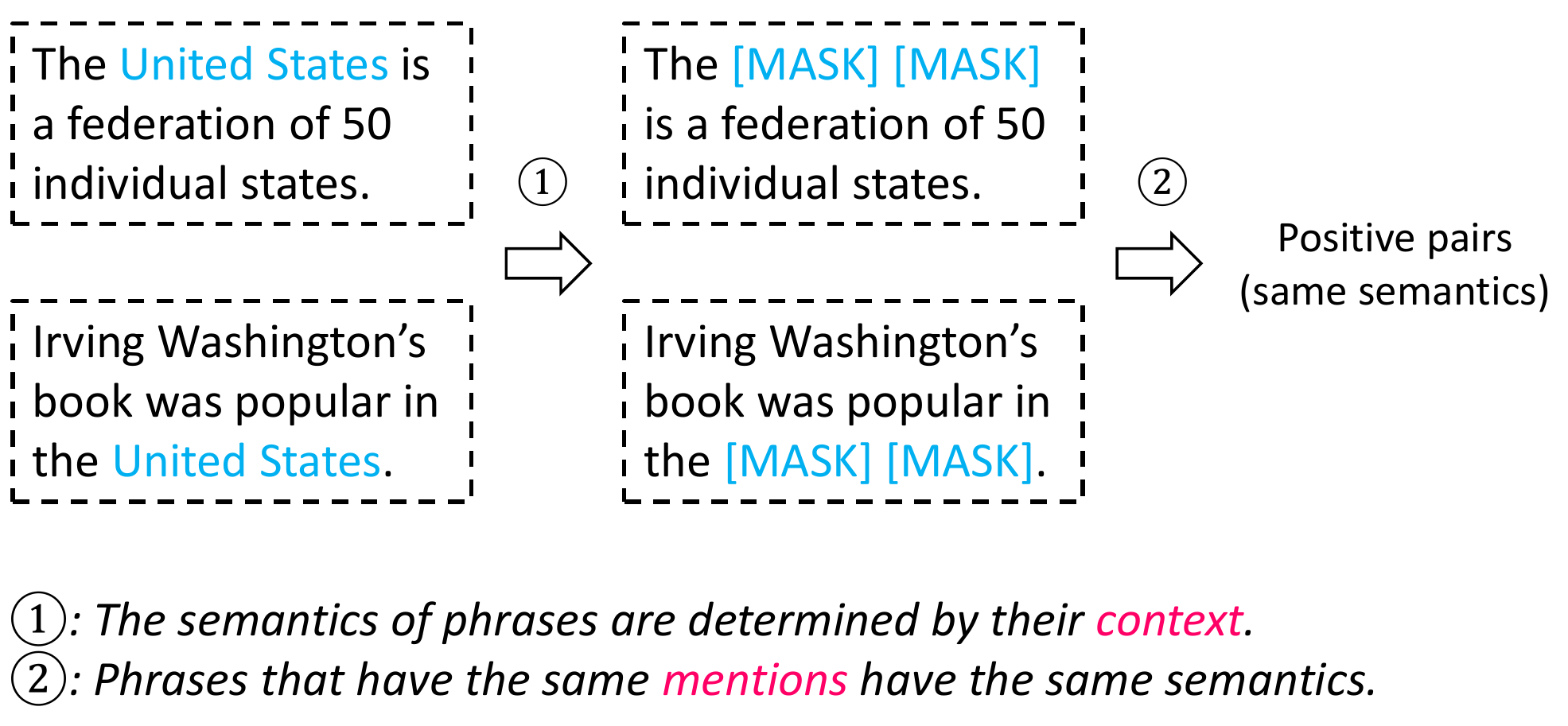}
    \caption{Two assumptions used in \our to produce positive pairs for contrastive learning.
    }
    \label{fig:hypos}
\end{figure}
To conduct contrastive learning for phrase representations, we first seek to produce contrastive pairs. Existing data augmentation methods for natural language processing (NLP) such as back translation~\citep{Xie2020UnsupervisedDA}, synonym replacement~\cite{Zhang2015CharacterlevelCN} and text mix up~\citep{Zhang2018mixupBE} are not designed for phrase-oriented noise, and thus cannot produce training pairs for phrase representation learning. In \our, we propose two assumptions about phrase semantics to obtain contrastive pairs:
\begin{enumerate}[leftmargin=*,nosep]
\item The phrase semantics are determined by their context.
\item Phrases that have the same mentions have the same semantics.
\end{enumerate}
As shown in Figure~\ref{fig:hypos}, given two sentences that contain the same phrase mentions (e.g.,~United States), we can mask the phrase mentions and the phrase semantics should
stay
the same based on assumption (1). Then, the phrase semantics from the two sentences are same as each other given assumption (2). Therefore, we can use the two masked sentences as positive pairs in  contrastive learning. The intuition behind the two assumptions is that we expect the phrase representations from different sentences describing the same phrase should group together in the latent space. 
Masking the phrase mentions forces the model to learn representations from context which prevents overfitting and representation collapse~\cite{Gao2021SimCSESC}.
Based on the two assumptions, our context-aware phrase representations can be pre-trained on a large corpus via a contrastive objective without supervision.

For
large-scale pre-training, we follow previous works~\citep{Chen2017OnSS, Henderson2017EfficientNL, Gao2021SimCSESC} and adopt in-batch negatives for training. However, we find 
in-batch negatives undermine the representation performance as finetuning (see~\Cref{tab:main_result}).
Because the number of topics is usually small in the finetuning dataset, examples in the same batch are likely to have the same topic. Hence, we cannot use in-batch negatives for data-specific finetuning. To solve this problem, we propose cluster-assisted contrastive learning (CCL) which leverages clustering results as pseudo-labels and sample negatives from highly confident examples in clusters. Cluster-assisted negative sampling has two advantages:
\begin{inparaenum}[(1)]
\item reducing potential positives from negative sampling compared to in-batch negatives;
\item the clusters are viewed as topics in documents, thus, cluster-assisted contrastive learning is a topic-specific finetuning process which pushes away instances from different topics in the latent space.
\end{inparaenum}

Based on the two assumptions and cluster-assisted negative sampling introduced in this paper, we pre-train phrase representations on a large-scale dataset and then finetune on a specific dataset for topic mining in an unsupervised way. In our experiments, we select LUKE~\cite{Yamada2020LUKEDC} as our backbone phrase representation model and pre-train it on Wikipedia~\footnote{https://dumps.wikimedia.org/} English corpus. To evaluate the quality of phrase representations, we conduct entity clustering on four datasets and find that pre-trained \our achieves $53.1\%$ (NMI) improvement compared to LUKE. After learning data-specific features with CCL, \our outperforms LUKE by $73.2\%$ (NMI) in average. We perform topical phrase mining on three datasets and comprehensive evaluation indicates \our extracts coherent and diverse topical phrases. Overall, our contributions are three-fold:
\begin{itemize}[leftmargin=*,nosep]
    \item We propose \our which produces superior phrase representations by unsupervised contrastive learning based on positive pairs from our phrase-oriented assumptions.
    \item To finetune on topic mining datasets, we propose a cluster-assisted negative sampling method for contrastive learning. This method reduces false negative instances caused by in-batch negatives and further improves phrase representations for topics accordingly.
    \item We conduct extensive experiments on entity type clustering and topic mining. Objective metrics and a user study show that \our can largely improve the phrase representations, then extracts more coherent and diverse topical phrases than existing topic mining methods. 
\end{itemize}

\begin{figure*}
    \centering
    \includegraphics[width=\linewidth]{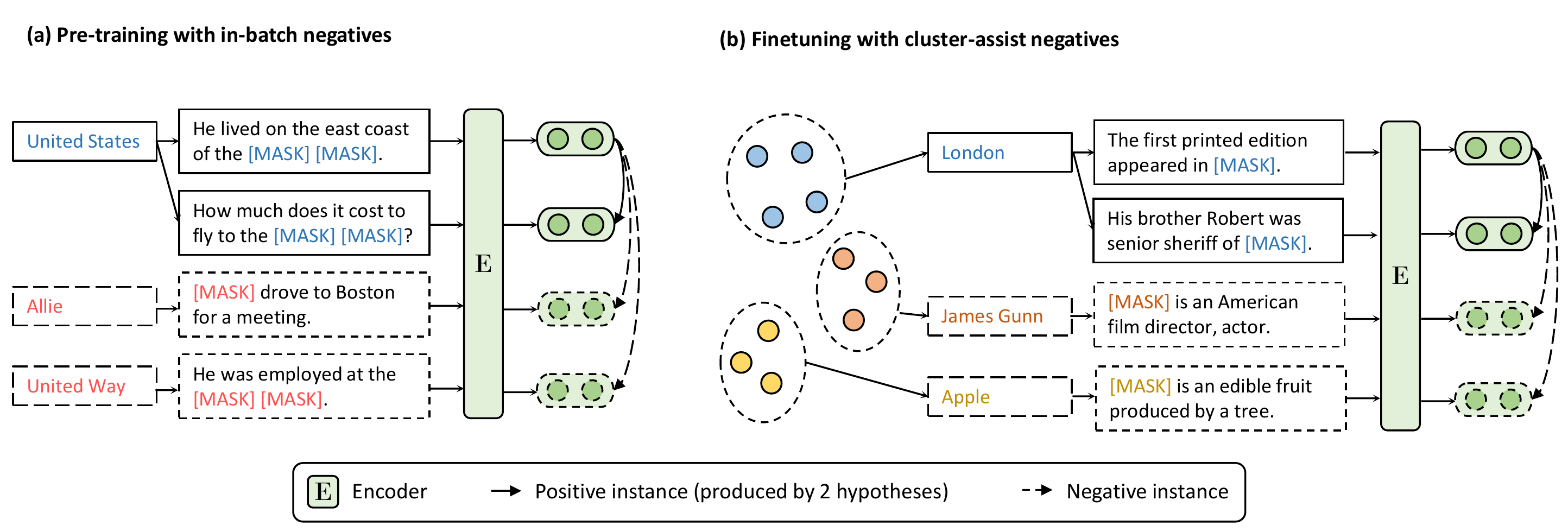}
    \caption{(a) Pre-training UCTopic on a  large-scale dataset with positive instances from our two assumptions and in-batch negatives. (b) Finetuning UCTopic on  a topic mining dataset with positive instances from our two assumptions and negatives from clustering.}
    \label{fig:framework}
\end{figure*}

\section{Background}
In this section, we introduce background knowledge about contrastive learning and our phrase encoder LUKE~\cite{Yamada2020LUKEDC}.

\subsection{Contrastive Learning}
Contrastive learning aims to learn effective representations by pulling semantically close neighbors together and pushing apart non-neighbors in the latent space~\cite{Hadsell2006DimensionalityRB}. Assume that we have a contrastive instance $\{x, x^{+}, x_1^{-},\dots,x_{N-1}^{-}\}$ including one positive and $N-1$ negative instances and their representations $\{\mathbf{h}, \mathbf{h}^{+}, \mathbf{h}_1^{-}, \dots, \mathbf{h}_{N-1}^{-}\}$ from the encoder, we follow the contrastive learning framework~\cite{Sohn2016ImprovedDM, Chen2020ASF, Gao2021SimCSESC} and take cross-entropy as our objective function:
\begin{equation}
    l = -\log \frac{e^{\mathrm{sim}(\mathbf{h}, \mathbf{h}^{+})/\tau}}{e^{\mathrm{sim}(\mathbf{h}, \mathbf{h}^{+})/\tau}+ \sum_{i=1}^{N-1}e^{\mathrm{sim}(\mathbf{h}, \mathbf{h}_i^{-})/\tau}}
\end{equation}
where $\tau$ is a temperature hyperparameter and $\mathrm{sim}(\mathbf{h}_1, \mathbf{h}_2)$ is the cosine similarity $\frac{\mathbf{h}_1^{\top}\mathbf{h}_2}{\Vert \mathbf{h}_1 \Vert \cdot \Vert \mathbf{h}_2 \Vert}$.

\subsection{Phrase Encoder}
In this paper, our phrase encoder $\mathbf{E}$ is transformer-based model LUKE~\cite{Yamada2020LUKEDC}. LUKE is a pre-trained language model that can directly output the representations of tokens and spans in sentences. Our phrase instance $x = (s, [l, r])$ includes a sentence $s$ and a character-level span $[l, r]$ ($l$ and $r$ are left and right boundaries of a phrase). $\mathbf{E}$ encodes the phrase $x$ and output the phrase representation $\mathbf{h} = \mathbf{E}(x) = \mathbf{E}(s, [l, r])$. Although LUKE can output span representations directly, we will show that span representations from LUKE are not able to represent phrases well (see~\Cref{sec:entity}). Different from LUKE, 
which is trained by predicting entities, \our is trained by contrastive learning on phrase contexts. Hence, the phrase presentations from \our are context-aware and robust to different domains.

\section{UCTopic}
\our is an unsupervised contrastive learning method for phrase representations and topic mining. Our goal is to learn a phrase encoder as well as  topic representations, so we can represent phrases effectively 
for general settings
and find topics from documents in an unsupervised way.
In this section, we introduce \our from two aspects: 
\begin{inparaenum}[(1)]
\item constructing positive pairs for phrases;
\item cluster-assisted contrastive learning.
\end{inparaenum}

\subsection{Positive Instances}
\label{sec:pos_ins}
One critical problem in constrastive learning is to how to construct positive pairs $(x, x^{+})$. Previous works~\cite{Wu2020CLEARCL, Meng2021COCOLMCA} apply augmentation techniques such as word deletion, reordering, and paraphrasing. However, these methods are not suitable for phrase representation learning. In this paper, we utilize the proposed assumptions introduced in~\Cref{intro} to construct positive instances for contrastive learning. 

Consider an example to understand our positive instance generation process: In~\Cref{fig:framework} (a), phrase \texttt{United States} appears in two different sentences ``\textit{He lived on the east coast of the \underline{United States}}'' and ``\textit{How much does it cost to fly to the \underline{United States}}''. We expect the phrase (\texttt{United States}) representations from the two sentences to be similar to reflect phrase semantics.
To encourage the model to learn phrase semantics from context and prevent the model from comparing phrase mentions in contrastive learning, we mask the phrase mentions with \texttt{[MASK]} token. The two masked sentences are used as positive instances. To decrease the inconsistency caused by masking between training and evaluation, in a positive pair, we keep one phrase mention unchanged in probability $p$.

Formally, suppose we have phrase instance $x=(s, [l,r])$ and its positive instance $x^{+}=(s', [l',r'])$ where $s$ denotes the sentence and $[l,r]$ are left and right boundaries of a phrase in $s$, we obtain the phrase representations $\textbf{h}$ and $\textbf{h}^{+}$ by encoder $\textbf{E}$ and apply in-batch negatives for pre-training. The training objective of \our becomes:
\begin{equation}
    l = -\log \frac{e^{\mathrm{sim}(\mathbf{h}, \mathbf{h}^{+})/\tau}}{\sum_{i=1}^{N}e^{\mathrm{sim}(\mathbf{h}, \mathbf{h}_i)/\tau}},
\end{equation}
for a mini-batch of $N$ instances, where $\mathbf{h}_i$ is an instance in a batch.

\subsection{Cluster-Assisted Contrastive Learning}
We find that contrastive learning with in-batch negatives on small datasets can undermine the phrase representations (see~\Cref{sec:entity}). 
Different from pre-training on a large corpus, in-batch negatives usually contain instances that have similar semantics as positives. For example, one document has three topics and our batch size is $32$. Thus, some instances in one batch are from the same topic but in-batch method views these instances as negatives with each other.
In this case, contrastive learning has noisy training signals and then results in decreasing performance.

To reduce the noise in negatives 
while optimizing
phrase representations according to topics in documents, we propose cluster-assisted contrastive learning (CCL). The basic idea is to utilize prior knowledge from pre-trained representations and clustering to reduce the noise existing in the negatives. Specifically, we first find the topics in documents with a clustering algorithm based on pre-trained phrase representations from \our. The centroids of clusters are considered as topic representations for phrases. After computing the cosine distance between phrase instances and centroids, we select $t$ 
percent of
instances that are close to centroids and assign pseudo labels to them. Then, the label of a phrase mention $p^m$~\footnote{phrase mentions are extracted from sentence $s$, i.e.,~$p^m=s[l:r]$} is determined by the majority vote of instances $\{x^m_0, x^m_1, \dots, x^m_n\}$ that contain $p^m$, where $n$ is the number of sentences assigned pseudo labels. In this way, we get some prior knowledge of phrase mentions for the following contrastive learning. 
See~\Cref{fig:framework} (b); three phrase mentions (\texttt{London}, \texttt{James Gunn} and \texttt{Apple}) which belong to three different clusters are labeled by different topic categories.

Suppose we have a topic set $\mathcal{C}$ in our documents, with phrases and their pseudo labels, we construct positive pairs $(x_{c_i}, x^+_{c_i})$ by method introduced in~\Cref{sec:pos_ins} for topic $c_i$ where $c_i \in \mathcal{C}$. 
To have contrastive instances, we randomly select phrases $p^m_{c_j}$ and instances $x^m_{c_j}$ from topic $c_j$ as negative instances $x^-_{c_j}$ in contrastive learning, where $c_j \in \mathcal{C} \wedge c_j \neq c_i$. As shown in~\Cref{fig:framework} (b), we construct positive pairs for phrase \texttt{London}, and use two phrases \texttt{James Gunn} and \texttt{Apple} from the other two clusters to randomly select negative instances. With pseudo labels, our method can avoid instances that have similar semantics as \texttt{London}. The training objective of finetuning is:
\begin{equation}
    l = -\log \frac{e^{\mathrm{sim}(\mathbf{h}_{c_i}, \mathbf{h}^{+}_{c_i})/\tau}}{e^{\mathrm{sim}(\mathbf{h}_{c_i}, \mathbf{h}^{+}_{c_i})/\tau}+ \sum_{c_j\in \mathcal{C}}e^{\mathrm{sim}(\mathbf{h}_{c_i}, \mathbf{h}_{c_j}^{-})/\tau}}.
\end{equation}
As for the masking strategy in pre-training, we conduct masking for all training instances but keep $x^+_{c_i}$ and $x^-_{c_j}$ unchanged in probability $p$.

To infer the topic $y$ of phrase instance $x$, we compute the cosine similarity between phrase representation $\mathbf{h}$ and topic representations $\mathbf{\tilde{h}}_{c_i}, c_i \in \mathcal{C}$. The nearest neighbor topic of $x$ is used as phrase topic. Formally,
\begin{equation}
    y = \mathrm{argmax}_{c_i \in \mathcal{C}}(\mathrm{sim}(\mathbf{h}, \mathbf{\tilde{h}}_{c_i}))
\end{equation}

\section{Experiments}
In this section, we evaluate the effectiveness of \our pre-training by contrastive learning. We start with entity clustering to compare the phrase representations from different methods. For topic modeling, we evaluate the topical phrases from three aspects and compare \our to other topic modeling baselines.
\subsection{Implementation Details}
To generate the training corpus, we use English Wikipedia~\footnote{https://en.wikipedia.org/} and extract text with hyper links as phrases. Phrases have the same entity ids from Wikidata~\footnote{https://www.wikidata.org/} or have the same mentions are considered as the same phrases (i.e.,~phrases have the same semantics).
We enumerate all sentence pairs containing the same phrase as positive pairs in contrastive learning. After processing, the pre-training dataset has $11.6$ million sentences and $108.8$ million training instances.

For pre-training, we start from a pretrained LUKE-BASE model~\cite{Yamada2020LUKEDC}. We follow previous works~\cite{Gao2021SimCSESC, BaldiniSoares2019MatchingTB} and two losses are used concurrently: the masked language model loss and the contrastive learning loss with in-batch negatives.  Our pre-training learning rate is 5e-5, batch size is $100$ and our model is optimized by AdamW in $1$ epoch. The probability $p$ of keeping phrase mentions unchanged is $0.5$ and the temperature $\tau$ in the contrastive loss is set to $0.05$.

\subsection{Entity Clustering}
\label{sec:entity}
To test the performance of phrase representations under objective tasks and metrics, we first apply \our on entity clustering and compare to other representation learning methods.

\noindent\textbf{Datasets}. We conduct entity clustering on four 
datasets with annotated entities
and their semantic categories are from general, review and biomedical domains:
\begin{inparaenum}[(1)]
\item CoNLL2003~\cite{Sang2003IntroductionTT} consists of 20,744 sentences extracted from Reuters news articles. We use Person, Location, and Organization entities in our experiments.\footnote{
We do not evaluate on the Misc category because it does not represent a single semantic category.
}
\item BC5CDR~\cite{Li2016BioCreativeVC} is the BioCreative V CDR task corpus. It contains 18,307 sentences from PubMed articles, with 15,953 chemical and 13,318 disease entities.
\item MIT Movie (MIT-M)~\cite{Liu2013QueryUE} contains 12,218 sentences with Title and Person entities.
\item W-NUT 2017~\cite{derczynski-etal-2017-results} focuses on identifying unusual entities in the context of emerging discussions and contains 5,690 sentences and six kinds of entities~\footnote{corporation, creative work, group, location, person, product}.
\end{inparaenum}

\label{sec:finetune}
\noindent\textbf{Finetuning Setup}. The learning rate for finetuning is 1e-5. We select $t$ (percent of instances) from $\{5, 10, 20, 50\}$. The probability $p$ of keeping phrase mentions unchanged and temperature $\tau$ in contrastive loss are the same as in pre-training settings. We apply K-Means to get pseudo labels for all experiments. Because \our is an unsupervised method, we use all data to finetune and evaluate. All results for finetuning are the best results during training process. We follow previous clustering works~\cite{Xu2017SelfTaughtCN, Zhang2021SupportingCW} and adopt Accuracy (ACC) and Normalized Mutual Information (NMI) to evaluate different approaches.

\begin{table*}[t]
\centering
\small
\begin{tabular}{rcccccccc}
\toprule
\multicolumn{1}{c|}{Datasets}        & \multicolumn{2}{c|}{\textbf{CoNLL2003}}   & \multicolumn{2}{c|}{\textbf{BC5CDR}}      & \multicolumn{2}{c|}{\textbf{MIT-M}}       & \multicolumn{2}{c}{\textbf{W-NUT2017}} \\ \midrule
\multicolumn{1}{c|}{Metrics}         & ACC            & \multicolumn{1}{c|}{NMI} & ACC            & \multicolumn{1}{c|}{NMI} & ACC            & \multicolumn{1}{c|}{NMI} & ACC               & NMI               \\ \midrule
\multicolumn{9}{c}{\textit{Pre-trained Representations}}                                                                                                                                                         \\ \midrule
\multicolumn{1}{r|}{Glove}           & 0.528          & 0.166                    & 0.587          & 0.026                    & 0.880          & 0.434                    & 0.368             & 0.188             \\
\multicolumn{1}{r|}{BERT-Ave.}       & 0.421          & 0.021                    & 0.857          & 0.489                    & 0.826          & 0.371                    & 0.270             & 0.034             \\
\multicolumn{1}{r|}{BERT-Mask}       & 0.430          & 0.022                    & 0.551          & 0.001                    & 0.587          & 0.001                    & 0.279             & 0.020             \\
\multicolumn{1}{r|}{LUKE}            & 0.590          & 0.281                    & 0.794          & 0.411                    & 0.831          & 0.432                    & 0.434             & 0.205             \\
\multicolumn{1}{r|}{DensePhrase}     & 0.603          & 0.172                    & 0.936          & 0.657                    & 0.716          & 0.293                    & 0.413             & 0.214             \\
\multicolumn{1}{r|}{Phrase-BERT}     & 0.643          & 0.297                    & 0.918          & 0.617                    & {\ul 0.916}    & {\ul 0.575}              & 0.452             & 0.241             \\
\multicolumn{1}{r|}{Ours w/o CCL} & {\ul 0.704}    & {\ul 0.464}              & {\ul 0.977}    & {\ul 0.846}              & 0.845          & 0.439                    & {\ul 0.509}       & {\ul 0.287}       \\ \midrule
\multicolumn{9}{c}{\textit{Finetuning on Pre-trained \our Representations}}                                                                                                                                   \\ \midrule
\multicolumn{1}{r|}{Ours w/ Class.}      & 0.703          & 0.458                    & 0.972          & 0.827                    & 0.738          & 0.323                    & 0.482             & 0.283             \\
\multicolumn{1}{r|}{Ours w/ In-B.}     & 0.706          & 0.470           & 0.974          & 0.834                    & 0.748          & 0.334                    & 0.454             & 0.301             \\
\multicolumn{1}{r|}{Ours w/ Auto.}     & 0.717          & 0.492           & 0.979          & 0.857                    & 0.858          & 0.458                    & 0.402             & 0.282             \\
\multicolumn{1}{r|}{\our}         & \textbf{0.743} & \textbf{0.495}                    & \textbf{0.981} & \textbf{0.865}           & \textbf{0.942} & \textbf{0.661}           & \textbf{0.521}    & \textbf{0.314}    \\ \bottomrule
\end{tabular}
\caption{Performance of entity clustering on four datasets from different domains. \emph{Class.}~represents using a classifier on pseudo labels. \emph{Auto.}~represents Autoencoder. The best results among all methods are bolded and the best results of pre-trained representations are underlined. \emph{In-B.}~represents contrastive learning with in-batch negatives.}
\label{tab:main_result}
\end{table*}
\noindent\textbf{Compared Baseline Methods}.
To demonstrate the effectiveness of our pre-training method and finetuning with cluster-assisted contrastive learning (CCL), we compare baseline methods from two aspects:

\noindent(1) Pre-trained token or phrase representations:
\begin{itemize}
    \item \textbf{Glove}~\cite{Pennington2014GloVeGV}. Pre-trained word embeddings on 6B tokens and dimension is $300$. We use averaging word embeddings as the representations of phrases.
    \item \textbf{BERT}~\cite{Devlin2019BERTPO}. Obtains phrase representations by averaging token representations (BERT-Ave.) or following CGExpan~\cite{Zhang2020EmpowerES} to substitute phrases with the \texttt{[MASK]} token, and use \texttt{[MASK]} representations as phrase embeddings (BERT-MASK).
    \item \textbf{LUKE}~\cite{Yamada2020LUKEDC}. Use as backbone model to show the effectiveness of our contrastive learning for pre-training and finetuning.
    \item \textbf{DensePhrase}~\cite{Lee2021LearningDR}. Pre-trained phrase representation learning in a supervised way for question answering problem. We use a pre-trained model released from the authors to get phrase representations.
    \item \textbf{Phrase-BERT}~\cite{Wang2021PhraseBERTIP}. Context-agnostic phrase representations from pretraining. We use a pre-trained model from the authors and get representations by phrase mentions.
    \item \textbf{Ours w/o CCL}. Pre-trained phrase representations of \our without cluster-assisted contrastive finetuning.
\end{itemize}

\noindent(2) Fine-tuning methods based on pre-trained representations of \our.
\begin{itemize}
    \item \textbf{Classifier}. We use pseudo labels as supervision to train a MLP layer and obtain a classifier of phrase categories.
    \item \textbf{In-Batch Contrastive Learning}. Same as contrastive learning for pre-training which uses in-batch negatives.
    \item \textbf{Autoencoder}. Widely used in previous neural topic and aspect extraction models~\cite{He2017AnUN, Iyyer2016FeudingFA, Tulkens2020EmbarrassinglySU}. We follow ABAE~\cite{He2017AnUN} to implement our autoencoder model for phrases.
\end{itemize}

\noindent\textbf{Experimental Results.} We report evaluation results of entity clustering in~\Cref{tab:main_result}. Overall, \our achieves the best results on all datasets and metrics. Specifically, \our improves the state-of-the-art method (Phrase-BERT) by $38.2\%$ NMI in average, and outperforms our backbone model (LUKE) by $73.2\%$ NMI.

When we compare different pre-trained representations, we find that our method (Ours w/o CCL) outperforms the other baselines on three datasets except MIT-M. There are two reasons:
\begin{inparaenum}[(1)]
\item All words in MIT-M dataset are lower case which is inconsistent with our pretraining dataset. The inconsistency between training and test causes performance to decay.
\item Sentences from MIT-M are usually short ($10.16$ words in average) compared to other datasets (e.g.,~$17.9$ words in W-NUT2017). Hence, \our can obtain limited contextual information with short sentences.
\end{inparaenum}
However, the performance decay caused by the two reasons can be eliminated by our CCL finetuning on datasets since on MIT-M \our achieves better results (0.661 NMI) than Phrase-BERT (0.575 NMI) after CCL.

On the other hand, compared to other finetuning methods, our CCL finetuning can further improve the pre-trained phrase representations by capturing data-specific features. The improvement is up to $50\%$ NMI on the MIT-M dataset. Ours w/ Class.~performs worse than our pre-trained \our in most cases which indicates that pseudo labels from clustering are noisy and cannot directly be used as supervision for representation learning. 
Ours w/ In-B.~is similar as Ours w/ Class.~which verifies our motivation on using CCL instead of in-batch negatives.
An autoencoder can improve pre-trained representations on three datasets but the margins are limited and the performance even drops on W-NUT2017. Compared to other finetuning methods, our CCL finetuning consistently improves  pre-trained phrase representations on different domains.

\begin{table}[t]
\small
\centering
\begin{tabular}{@{}r|cccc@{}}
\toprule
\multicolumn{1}{c|}{Model}  & \multicolumn{2}{c|}{\textbf{UCTopic}}                                                                             & \multicolumn{2}{c}{\textbf{LUKE}}                                                                                 \\ \midrule
\multicolumn{1}{c|}{Metric} & ACC                                                     & \multicolumn{1}{c|}{NMI}                                & ACC                                                     & NMI                                                     \\ \midrule
Context+Mention             & 0.44                                                    & 0.29                                                    & 0.39                                                    & 0.21                                                    \\ \midrule
Mention                     & \begin{tabular}[c]{@{}c@{}}0.32 \\ (-27\%)\end{tabular} & \begin{tabular}[c]{@{}c@{}}0.15\\  (-48\%)\end{tabular} & \begin{tabular}[c]{@{}c@{}}0.28 \\ (-28\%)\end{tabular} & \begin{tabular}[c]{@{}c@{}}0.10 \\ (-52\%)\end{tabular} \\ \midrule
Context                     & \begin{tabular}[c]{@{}c@{}}0.43 \\ (-3\%)\end{tabular}  & \begin{tabular}[c]{@{}c@{}}0.16 \\ (-44\%)\end{tabular} & \begin{tabular}[c]{@{}c@{}}0.27\\ (-31\%)\end{tabular}  & \begin{tabular}[c]{@{}c@{}}0.07 \\ (-67\%)\end{tabular} \\ \bottomrule
\end{tabular}
\caption{Ablation study on the input of phrase instances of W-NUT 2017. \our here is pre-trained representations without CCL finetuning. Percentages in brackets are changes compared to Context+Mention.}
\label{tab:c_or_m}
\end{table}

\noindent\textbf{Context or Mentions}.
To investigate the source of \our phrase semantics (i.e.,~phrase mentions or context), 
we conduct an ablation study on the type of input and compare \our to LUKE. To eliminate the influence of repeated phrase mentions on clustering results, we use only one phrase instance (i.e.,~sentence and position of a phrase) for each phrase mention. As shown in~\Cref{tab:c_or_m}, there are three types of inputs:
\begin{inparaenum}[(1)]
\item Context+Mention: The same input as experiments in~\Cref{tab:main_result} including the whole sentence that contains the phrase.
\item Mention: Use only phrase mentions as inputs of the two models.
\item Context: We mask the phrase mentions in sentences and models can only get information from the context.
\end{inparaenum}
We can see that \our gets more information from context ($0.43$ ACC, $0.16$ NMI) than mentions ($0.32$ ACC, $0.15$ NMI). Compared to LUKE, \our is more robust to phrase mentions (when predicting on only context, \our $-3\%$ ACC and $-44\%$ NMI vs.~LUKE $-31\%$ ACC and $-67\%$ NMI).

\subsection{Topical Phrase Mining}
\label{sec:topic}
In this section, we apply \our on topical phrase mining and conduct human evaluation to show our model outperforms previous topic model baselines.

\noindent\textbf{Experiment Setup}. 
To find topical phrases in documents, we first extract noun phrases by spaCy~\footnote{https://spacy.io/} noun chunks and remove single pronoun words. Before CCL finetuning, we obtain the number of topics for each dataset by computing the Silhouette Coefficient~\cite{Rousseeuw1987SilhouettesAG}.

{
Specifically, we randomly sample 10K phrases from the dataset and apply K-Means clustering on pre-trained \our phrase representations with different numbers of cluster. We compute Silhouette Coefficient scores for different topic numbers; the number with the largest score will be used as the topic number in a dataset.
}
Then, we conduct CCL on the dataset with the same settings as described in~\Cref{sec:finetune}. Finally, after obtaining topic distribution $\mathbf{z}_x \in \mathbb{R}^{|\mathcal{C}|}$ for a phrase instance $x$ in a sentence, we get context-agnostic phrase topics by using averaged topic distribution $\mathbf{z}_{p^m} = \frac{1}{n}\sum_{1 \leq i \leq n}\mathbf{z}_{x_i^m}$, where phrase instances $\{x_i^m\}$ in different sentences have the same phrase mention $p^m$. The topic of a phrase mention has the highest probability in $\mathbf{z}_{p^m}$.

\noindent\textbf{Dataset}.
We conduct topical phrase mining on three datasets from news, review and computer science domains.
\begin{itemize}
    \item \textbf{Gest}. We collect restaurant reviews from Google Local\footnote{https://www.google.com/maps} and use 100K reviews containing 143,969 sentences for topical phrase mining.
    \item \textbf{KP20k}~\cite{Meng2017DeepKG} is a collection of titles and abstracts from computer science papers. 500K sentences are used in our experiments.
    \item \textbf{KPTimes}~\cite{Gallina2019KPTimesAL} includes news articles from the New York Times from 2006 to 2017 and 10K news articles from the Japan Times. We use 500K sentences for topical phrase mining.
\end{itemize}

\begin{table}[t]
\centering
\small
\begin{tabular}{@{}r|ccc@{}}
\toprule
Datasets     & \textbf{Gest} & \textbf{KP20k} & \textbf{KPTimes} \\ \midrule
\# of topics & 22   & 10    & 16      \\ \bottomrule
\end{tabular}
\caption{The numbers of topics in three datasets.}
\label{tab:topics}
\end{table}
The number of topics determined by Silhouette Coefficient is shown in~\Cref{tab:topics}.

\noindent\textbf{Compared Baseline Methods}.
We compare \our against three topic baselines:
\begin{itemize}
    \item \textbf{Phrase-LDA}~\cite{pLDA}. LDA model incorporates phrases by simply converting phrases into unigrams (e.g.,~``city view'' to ``city\_view'').
    \item \textbf{TopMine}~\cite{ElKishky2014ScalableTP}. A scalable pipeline that partitions a document into phrases, then uses phrases as constraints to ensure all words are placed under the same topic.
    \item \textbf{PNTM}~\cite{Wang2021PhraseBERTIP}. A topic model with Phrase-BERT by using an autoencoder that reconstructs a document representation. The model is viewed as the state-of-the-art topic model.
\end{itemize}
We do not include topic models such as LDA~\cite{Blei2003LatentDA}, PD-LDA~\cite{Lindsey2012APT}, TNG~\cite{Wang2007TopicalNP}, KERT~\cite{Danilevsky2014AutomaticCA} as baselines, because these models are compared in TopMine and PNTM. For Phrase-LDA and PNTM, we use the same phrase list produced by \our. TopMine uses phrases produced by itself.

\begin{figure}
    \centering
    \includegraphics[width=\linewidth]{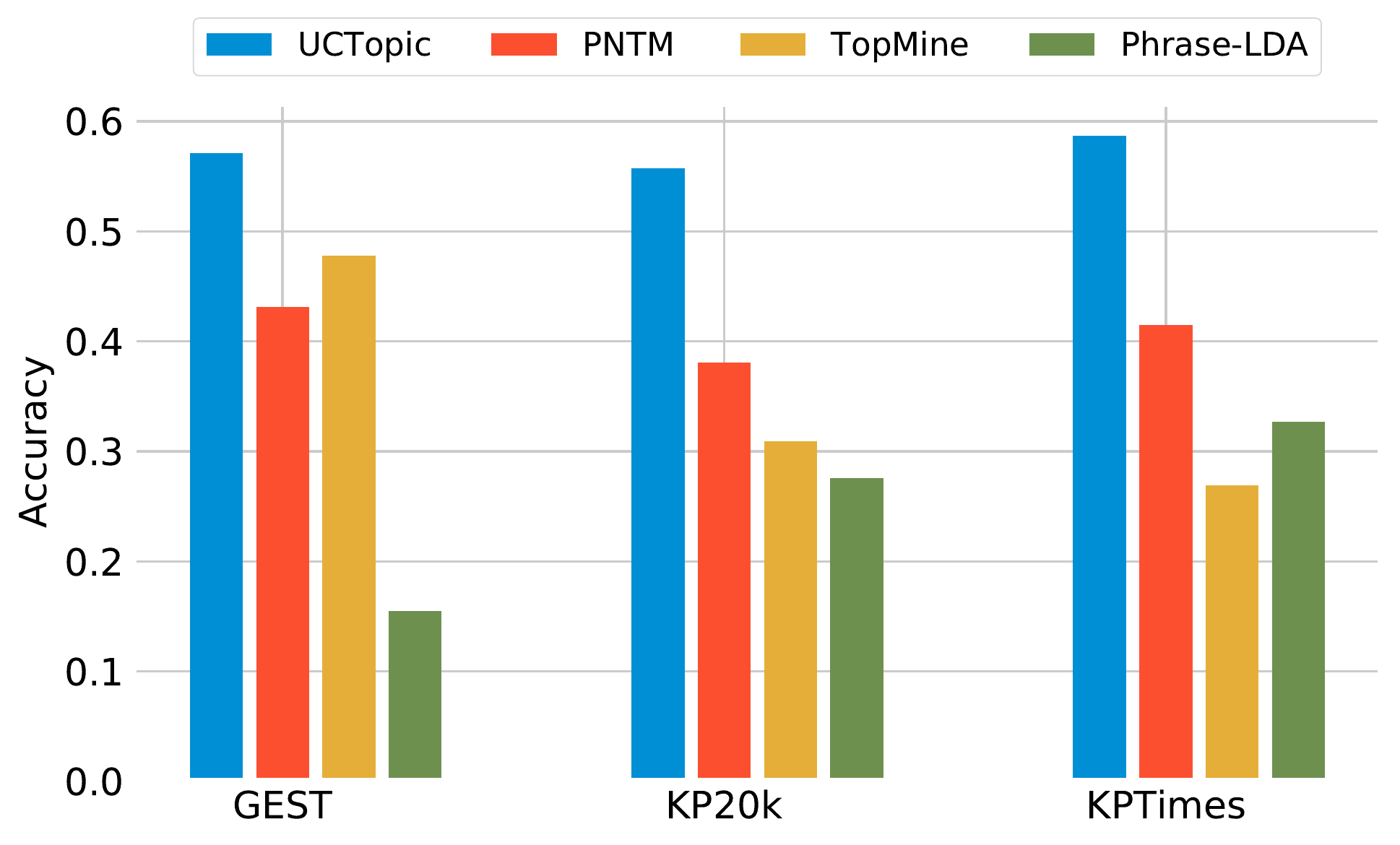}
    \caption{Results of phrase intrusion task.}
    \label{fig:intrusion}
\end{figure}

\begin{table}[t]
\small
\centering
\begin{tabular}{@{}l|cccc@{}}
\toprule
      & \multicolumn{1}{l}{\textbf{\our}} & \multicolumn{1}{l}{\textbf{PNTM}} & \multicolumn{1}{l}{\textbf{TopMine}} & \multicolumn{1}{l}{\textbf{P-LDA}} \\ \midrule
Gest  & 20                          & 18                       & 20                          & 11                              \\
KP20k & 10                           & 9                        & 9                           & 4                              \\ \bottomrule
\end{tabular}
\caption{Number of coherent topics on Gest and KP20k.}
\label{tab:num_topics}
\end{table}

\begin{figure}[t]
    \centering
    \includegraphics[width=\linewidth]{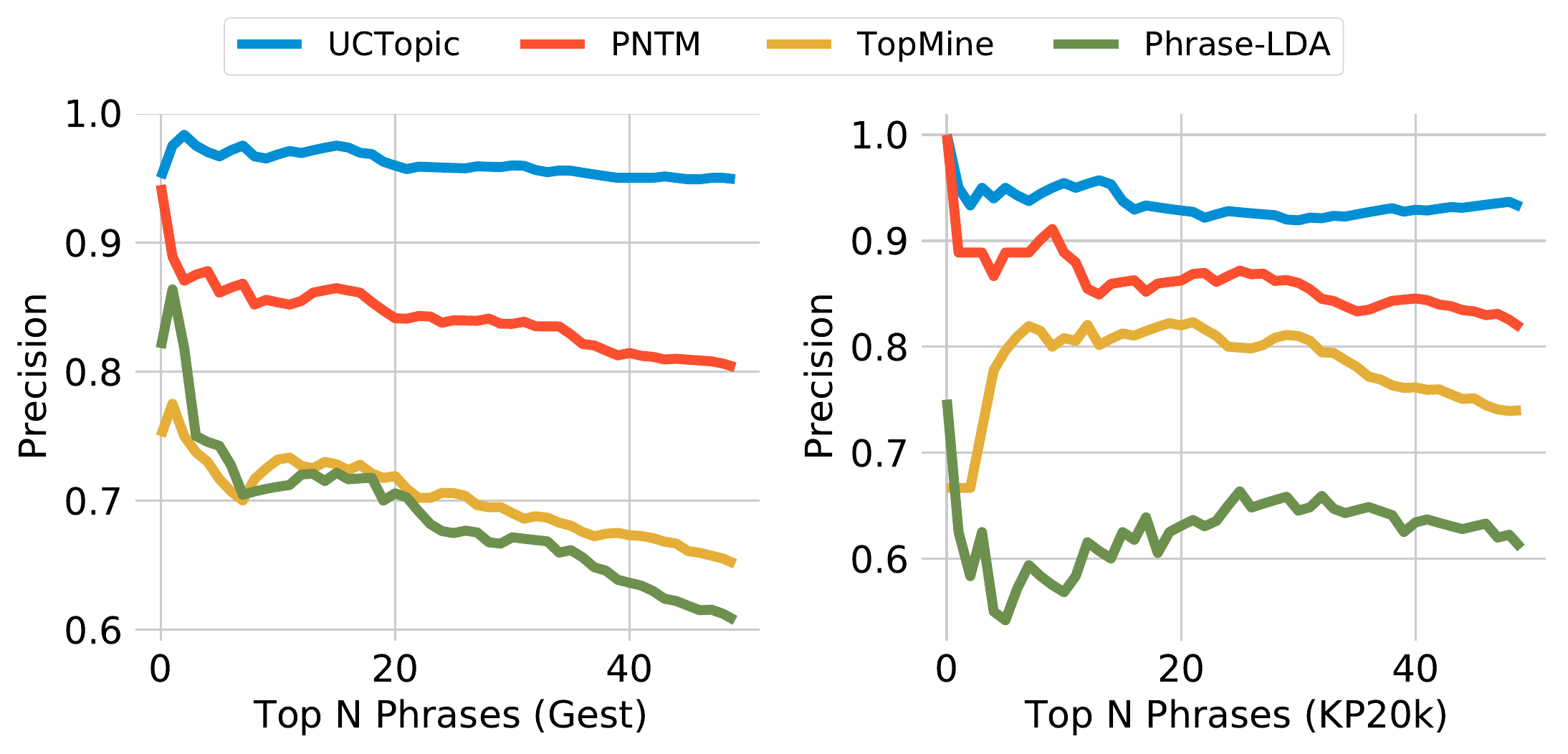}
    \caption{Results of top n precision.}
    \label{fig:precision}
\end{figure}

\begin{table}[t]
\centering
\small
\begin{tabular}{@{}l|cc|cc@{}}
\toprule
Datasets & \multicolumn{2}{c|}{\textbf{Gest}} & \multicolumn{2}{c}{\textbf{KP20k}} \\ \midrule
Metrics  & tf-idf           & word-div.       & tf-idf           & word-div.       \\ \midrule
TopMine  & \textbf{0.5379}  & 0.6101          & 0.2551  & 0.7288          \\
PNTM     & 0.5152           & 0.5744          & \textbf{0.3383}           & 0.6803          \\
UCTopic  & 0.5186           & \textbf{0.7486} & 0.3311           & \textbf{0.7600} \\ \bottomrule
\end{tabular}
\caption{Informativeness (tf-idf) and diversity (word-div.) of extracted topical phrases.}
\label{tab:div}
\end{table}

\begin{table*}[]
\centering
\scalebox{0.82}{
\small
\setlength{\tabcolsep}{1.2mm}{
\begin{tabular}{@{}lllll|ll@{}}
\toprule
\multicolumn{5}{c|}{\textbf{Gest}}                                                                                                                                                                   & \multicolumn{2}{c}{\textbf{KP20k}}                                        \\ \midrule
\multicolumn{2}{c|}{\textit{Drinks}}                                          & \multicolumn{3}{c|}{\textit{Dishes}}                                                                                 & \multicolumn{2}{c}{\textit{Programming}}                                  \\ \midrule
\multicolumn{1}{c|}{\our}        & \multicolumn{1}{c|}{PNTM}               & \multicolumn{1}{c|}{\our}                  & \multicolumn{1}{c|}{PNTM}             & \multicolumn{1}{c|}{TopMine} & \multicolumn{1}{c|}{\our}                & \multicolumn{1}{c}{TopMine} \\ \midrule
\multicolumn{1}{l|}{lager}          & \multicolumn{1}{l|}{drinks}             & \multicolumn{1}{l|}{cauliflower fried rice
}     & \multicolumn{1}{l|}{great burger}     & mac cheese                   & \multicolumn{1}{l|}{markup language}        & software development        \\
\multicolumn{1}{l|}{whisky}         & \multicolumn{1}{l|}{bar drink}          & \multicolumn{1}{l|}{chicken tortilla soup}    & \multicolumn{1}{l|}{great elk burger} & ice cream                    & \multicolumn{1}{l|}{scripting language}     & software engineering        \\
\multicolumn{1}{l|}{vodka}          & \multicolumn{1}{l|}{just drink}         & \multicolumn{1}{l|}{chicken burrito}          & \multicolumn{1}{l|}{great hamburger}  & potato salad                 & \multicolumn{1}{l|}{language construct}     & machine learning            \\
\multicolumn{1}{l|}{whiskey}        & \multicolumn{1}{l|}{alcohol}            & \multicolumn{1}{l|}{
fried calamari}             & \multicolumn{1}{l|}{good burger}      & french toast                 & \multicolumn{1}{l|}{java library}           & object oriented             \\
\multicolumn{1}{l|}{rum}            & \multicolumn{1}{l|}{liquor}             & \multicolumn{1}{l|}{roast beef sandwich}      & \multicolumn{1}{l|}{good hamburger}   & chicken sandwich             & \multicolumn{1}{l|}{programming structure}  & open source                 \\
\multicolumn{1}{l|}{own beer}       & \multicolumn{1}{l|}{booze}              & \multicolumn{1}{l|}{grill chicken sandwich}   & \multicolumn{1}{l|}{awesome steak}    & cream cheese                 & \multicolumn{1}{l|}{xml syntax}             & design process              \\
\multicolumn{1}{l|}{ale}            & \multicolumn{1}{l|}{drink order}        & \multicolumn{1}{l|}{buffalo chicken sandwich} & \multicolumn{1}{l|}{burger joint}     & fried chicken                & \multicolumn{1}{l|}{module language}        & design implementation       \\
\multicolumn{1}{l|}{craft cocktail} & \multicolumn{1}{l|}{ok drink}           & \multicolumn{1}{l|}{pull pork sandwich}       & \multicolumn{1}{l|}{woody 's bbq}     & fried rice                   & \multicolumn{1}{l|}{programming framework}  & programming language        \\
\multicolumn{1}{l|}{booze}          & \multicolumn{1}{l|}{alcoholic beverage} & \multicolumn{1}{l|}{chicken biscuit}          & \multicolumn{1}{l|}{excellent burger} & french fries                 & \multicolumn{1}{l|}{object-oriented language} & source code                 \\
\multicolumn{1}{l|}{tap beer}       & \multicolumn{1}{l|}{beverage}           & \multicolumn{1}{l|}{tortilla soup}            & \multicolumn{1}{l|}{beef burger}      & bread pudding                & \multicolumn{1}{l|}{python module}          & support vector machine      \\ \bottomrule
\end{tabular}}}
\label{tab:examples}
\caption{Top topical phrases on Gest and KP20k and the minimum phrase frequency is 3.}
\end{table*}

\noindent\textbf{Topical Phrase Evaluation}.
We evaluate the quality of topical phrases from three aspects:
\begin{inparaenum}[(1)]
\item \emph{topical separation};
\item \emph{phrase coherence};
\item \emph{phrase informativeness and diversity}.
\end{inparaenum}

To evaluate \emph{topical separation}, we perform the \textbf{phrase intrusion} task following previous work~\cite{ElKishky2014ScalableTP, Chang2009ReadingTL}. The phrase intrusion task involves a set of questions asking humans to discover the `intruder' phrase from other phrases.
{
In our experiments, each question has $6$ phrases and $5$ of them are randomly sampled from the top 50 phrases of one topic and the remaining phrase is randomly chosen from another topic (top 50 phrases). Annotators are asked to select the intruder phrase.
We sample $50$ questions for each method and each dataset ($600$ questions in total) and shuffle all questions. Because these questions are sampled independently, we asked $4$ annotators to answer these questions and each annotator answers $150$ questions on average. 
}
Results of the task evaluate how well the phrases are separated by topics.
The evaluation results are shown in~\Cref{fig:intrusion}. \our outperforms other baselines on three datasets, which means our model can find well-separated topics in documents.

To evaluate \emph{phrase coherence} in one topic, we follow ABAE~\cite{He2017AnUN} and ask annotators to evaluate if the top 50 phrases from one topic are coherent (i.e.,~most phrases represent the same topic). $3$ annotators evaluate four models on Gest and KP20k datasets. Numbers of coherent topics are shown in~\Cref{tab:num_topics}. We can see that \our, PNTM and TopMine can recognize similar numbers of coherent topics, but the numbers of Phrase-LDA are less than the other three models. For a coherent topic, each of the top phrases will be labeled as correct if the phrase reflects the related topic. Same as ABAE, we adopt \emph{precision@n} to evaluate the results. \Cref{fig:precision} shows the results; we can see that \our substantially outperforms other models. \our can maintain high precision with a large \emph{n} when the precision of other models decreases.

Finally, to evaluate \emph{phrase informativeness and diversity}, we use tf-idf and word diversity (word-div.) to evaluate the top topical phrases.
{
Basically, informative phrases cannot be very common phrases in a corpus (e.g.,~``good food'' in Gest) and we use tf-idf to evaluate the ``importance'' of a phrase. To eliminate the influence of phrase length, we use averaged word tf-idf in a phrase as the phrase tf-idf. Specifically, $\text{tf-idf}(p, d) = \frac{1}{m}\sum_{1\leq i\leq m}\text{tf-idf}(w_i^{p})$, where $d$ denotes the document and $p$ is the phrase. In our experiments, a document is a sentence in a review.

In addition, we hope that our phrases are diverse enough in a topic instead of expressing the same meaning (e.g.,~``good food'' and ``great food''). To evaluate the diversity of the top phrases, we calculate the ratio of distinct words among all words. Formally, given a list of phrases $[p_1, p_2,\dots,p_n]$, we tokenize the phrases into a word list $\mathbf{w} = [w_1^{p_1}, w_2^{p_1},\dots,w_m^{p_n}]$; $\mathbf{w}'$ is the set of unique words in $\mathbf{w}$. The word diversity is computed by $\frac{|\mathbf{w}'|}{|\mathbf{w}|}$. We only evaluate coherent topics labeled in \emph{phrase coherence}; the coherent topic numbers of Phrase-LDA are smaller than others, hence we evaluate the other three models. 
}

We compute the tf-idf and word-div.~on the top 10 phrases and use the averaged value on topics as final scores. 
Results are shown in~\cref{tab:div}. 
PNTM and \our achieve similar tf-idf scores, because the two methods use the same phrase lists extracted from spaCy.
\our extracts the most diverse phrases in a topic, because our phrase representations are more context-aware. In contrast, since PNTM gets representations dependent on phrase mentions, the phrases from PNTM contain the same words and hence are less diverse.

\noindent\textbf{Case Study}.
We compare top phrases from \our, PNTM and TopMine in~\Cref{tab:examples}. From examples, we can see the phrases are consistent with our user study and diversity evaluation. Although the phrases from PNTM are coherent, the diversity of phrases is less than others (e.g.,~``drinks'', ``bar drink'', ``just drink'' from Gest) because context-agnostic representations let similar phrase mentions group together. The phrases from TopMine are diverse but are not coherent in some cases (e.g.,~``machine learning'' and ``support vector machine'' in the programming topic). In contrast, \our can extract coherent and diverse topical phrases from documents.
\section{Related Work}
Many attempts 
have been made to extract topical phrases via LDA~\cite{Blei2003LatentDA}. \citet{Wallach2006TopicMB} incorporated a bigram language model into LDA by a
hierarchical dirichlet generative probabilistic model
to share the topic across each word within a bigram. TNG~\cite{Wang2007TopicalNP} applied additional latent variables and word-specific multinomials to model bi-grams and combined bi-grams to form n-gram phrases. PD-LDA~\cite{Lindsey2012APT} used a hierarchical Pitman-Yor process to share the same topic among all words in a given n-gram. \citet{Danilevsky2014AutomaticCA} ranked the resultant phrases based on four heuristic metrics. TOPMine~\cite{ElKishky2014ScalableTP} proposed to restrict all constituent terms within a phrase to share the same latent topic and assign a phrase to the topic of its constituent words. Compared to previous topic mining methods, \our 
builds on the
success of pre-trained language models and unsupervised contrastive learning on a large-scale dataset. Therefore, \our provides high-quality pre-trained phrase representations and state-of-the-art finetuning for topic mining.

Early works in phrase representation build upon a composition function that combines component word embeddings together into simple phrase embedding. \citet{Yu2015LearningCM} implemented the function by rule-based composition over word vectors. \citet{Zhou2017LearningPE} applied a pair-wise GRU model and datasets such as PPDB~\cite{Pavlick2015PPDB2B} to learn phrase representations. Phrase-BERT~\cite{Wang2021PhraseBERTIP} composed token embeddings from BERT and pretrained on positive instances produced by GPT-2-based diverse paraphrasing model~\cite{Krishna2020ReformulatingUS}. \citet{Lee2021LearningDR} learned phrase representations from the supervision of reading comprehension tasks and applied representations on open-domain QA. Other works learned phrase embeddings for specific tasks such as semantic parsing~\cite{Socher2011ParsingNS} and machine translation~\cite{Bing2015AbstractiveMS}. In this paper, we present unsupervised contrastive learning method for pre-training phrase representations of general purposes and for finetuning to topic-specific phrase representations.
\section{Conclusion}
In this paper, we propose \our, a contrastive learning framework that can effectively learn phrase representations
without supervision. To finetune on topic mining datasets, we propose cluster-assisted contrastive learning which reduces noise by selecting negatives from clusters. During finetuning, our phrase representations are optimized for topics in the document hence the representations are further improved. We conduct comprehensive experiments on entity clustering and topical phrase mining. Results show that \our largely improves phrase representations. Objective metrics and a user study indicate \our can extract coherent and diverse topical phrases.

\clearpage
\bibliography{custom}
\bibliographystyle{acl_natbib}



\end{document}